\documentclass[conference,compsoc]{IEEEtran}
\usepackage{tabularx}
\usepackage{booktabs}
\usepackage{multirow}
\usepackage{graphicx}

\usepackage{amsfonts}
\usepackage{color} 
\ifCLASSOPTIONcompsoc
  \usepackage[nocompress]{cite}
\else
  \usepackage{cite}
\fi
\ifCLASSINFOpdf
\else
\fi
\usepackage{amsmath,amsfonts}
\usepackage{amssymb}      
\usepackage{algorithm}    
\usepackage{algpseudocode} 
\algrenewcommand\algorithmicrequire{\textbf{Input:}}
\algrenewcommand\algorithmicensure{\textbf{Output:}}

\begin{document}
%
\title{Compressing Multi-Task Model for Autonomous Driving via Pruning and Knowledge Distillation}

\author{\IEEEauthorblockN{Jiayuan Wang}
\IEEEauthorblockA{
University of Windsor\\
401 Sunset Ave, Windsor, ON N9B 3P4\\
Email: wang621@uwindsor.ca}
\and
\IEEEauthorblockN{Q. M. Jonathan Wu}
\IEEEauthorblockA{
University of Windsor\\
401 Sunset Ave, Windsor, ON N9B 3P4\\
Email: jwu@uwindsor.ca}
\and
\IEEEauthorblockN{Ning Zhang}
\IEEEauthorblockA{
University of Windsor\\
401 Sunset Ave, Windsor, ON N9B 3P4\\
Email: nzhang@uwindsor.ca}
\and
\IEEEauthorblockN{Katsuya Suto}
\IEEEauthorblockA{
Hokkaido University\\
Email: k.suto@ist.hokudai.ac.jp}
\and
\IEEEauthorblockN{Lei Zhong}
\IEEEauthorblockA{Toyota Motor Corporation\\
Tokyo, 100-0004, Japan\\
Email: lei.zhong@toyota.global}
}


%


\maketitle

\begin{abstract}
Autonomous driving systems rely on panoptic perception to jointly handle object detection, drivable area segmentation, and lane line segmentation. Although multi-task learning is an effective way to integrate these tasks, its increasing model parameters and complexity make deployment on on-board devices difficult. To address this challenge, we propose a multi-task model compression framework that combines task-aware safe pruning with feature-level knowledge distillation. Our safe pruning strategy integrates Taylor-based channel importance with gradient conflict penalty to keep important channels while removing redundant and conflicting channels. To mitigate performance degradation after pruning, we further design a task head-agnostic distillation method that transfers intermediate backbone and encoder features from a teacher to a student model as guidance. Experiments on the BDD100K dataset demonstrate that our compressed model achieves a 32.7\% reduction in parameters while segmentation performance shows negligible accuracy loss and only a minor decrease in detection (-1.2\% for Recall and -1.8\% for mAP50) compared to the teacher. The compressed model still runs at 32.7 FPS in real-time. These results show that combining pruning and knowledge distillation provides an effective compression solution for multi-task panoptic perception.
\end{abstract}


%
\IEEEpeerreviewmaketitle

\section{Introduction}
With the rapid advancements in autonomous driving, extensive research \cite{zhan2024yolopx, pan2025gradient} reveals that the panoptic driving perception system is an important component for the ego-vehicle to understand the surroundings. In general, a panoptic driving perception system \cite{wu2022yolop} includes tasks of object detection, drivable area segmentation, and lane line segmentation. Object detection enables ego-vehicles to recognize obstacles to avoid collision. Drivable area segmentation and lane detection are important for planning safe and efficient driving routes. Recent studies \cite{wang2024you, wang2025rmt} demonstrate that a multi-task model can perform all these tasks simultaneously, making a unified perception pipeline that is both efficient and effective. Compared to implementing each task with an individual model, the multi-task model significantly reduces the computational cost. Specifically, by sharing a general encoder, the multi-task model avoids redundant calculations and can achieve real-time inference for all tasks, even in an on-board system with limited compute resources. 

Although multi-task learning is an effective way to integrate these tasks, the continued development of multi-task models leads to a rapid growth in model parameters and complexity. The models adopt deeper backbones, more channels, and specialized heads to improve accuracy. With the number of parameters and tasks growing, the requirements for memory and processing power are increasing \cite{deng2020model}. As a result, the advantage of multi-task efficiency is gradually offset. Deploying a large-scale model on resource-constrained on-board platforms has been a major challenge. Therefore, it is necessary to explore effective model compression techniques for multi-task models to ensure their deployability and applicability in real-world autonomous driving scenarios.

Model compression techniques aim to reduce the parameters and complexity of neural networks without significantly compromising their accuracy \cite{dantas2024comprehensive}. The common methods include pruning, quantization, low-rank factorization, and knowledge distillation. Pruning removes the model's redundant or less important weights and connections to reduce parameters and computation \cite{vadera2022methods}. Quantization reduces the precision of model weights and activations to save memory and accelerate inference \cite{fan2024multi}, such as switching from floating-point to lower-bit representations. Low-rank factorization decomposes large and dense weight matrices into smaller and lower-rank matrices to reduce parameters while maintaining representation with minimal performance loss \cite{swaminathan2020sparse}. Knowledge distillation transfers knowledge from a large and accurate teacher model to a smaller student model to maintain performance with lower complexity \cite{moslemi2024survey}. These model compression techniques have proven to be effective in single-task deep learning models. However, model compression for multi-task learning models is still under-explored. In multi-task learning, the shared parameters serve multiple tasks simultaneously, which means that the importance of a channel or weight cannot be judged by a single task loss alone. A feature that appears redundant for one task may be critical for another. Similarly, applying knowledge distillation in multi-task learning is challenging because the student must maintain performance across heterogeneous tasks rather than optimizing for a single task. Therefore, effective compression for multi-task learning must consider both channel importance and inter-task interference, which makes directly adopting traditional pruning and knowledge distillation strategies insufficient.

In this paper, we propose a multi-task model compression framework that integrates safe pruning with feature-level knowledge distillation for autonomous driving perception tasks. To ensure safe pruning in a multi-task learning setting, we first compute the channel importance for each task using a Taylor-based criterion. Then, incorporate an inter-task gradient conflict penalty. This strategy preserves channels that are essential for any task, while pruning those that are redundant or strongly conflicting. After pruning, we restore the student model’s performance using head-agnostic feature knowledge distillation. In this approach, the student’s backbone and encoder features are trained to mimic the corresponding features of the teacher model. This method simplifies the distillation process by avoiding task-specific head logits, while still transferring useful intermediate representations to recover overall performance. Experiments on the BDD100K \cite{yu2020bdd100k} dataset show that our compressed model achieves a parameter reduction of approximately 32.7\% while maintaining nearly the same performance across drivable area segmentation and lane line segmentation tasks as the teacher model. Only the detection task is minor decreased, but still acceptable.

\section{Proposed Methodology}
Our goal is to compress a multi-task model by removing unnecessary channels and then to recover its performance using knowledge distillation. Specifically, our framework follows a two-stage design. In the first stage, we perform multi-task safe pruning by estimating the channel importance for each task and also analyzing inter-task gradient conflicts. In the second stage, we apply feature-level knowledge distillation to transfer intermediate representations from the teacher to the pruned student model, allowing the student to recover performance across all tasks. 

\subsection{Pruning}
Our proposed safe pruning strategy consists of three components: Taylor-based channel importance estimation to quantify the contribution of each channel to each task. Inter-task gradient conflict penalty to identify channels that hinder multi-task optimization. A safe pruning mechanism that combines importance and gradient conflict information.

\subsubsection{Taylor-based Channel Importance} 
Inspired by \cite{molchanov2019importance}, we extend their proposed Taylor pruning methods to multi-task learning scenarios. Specifically, we quantify the importance of each channel in the network to each task using a first-order Taylor approximation of that task’s loss. Specifically, for a given task $t$, the importance score of channel $c$ is defined as
\begin{equation}
I_c^t = \mathbb{E}_{N,H,W}\left[ \, | a_c \cdot g_c | \, \right],
\end{equation}
where $a_c$ denotes the activation of channel $c$, and $g_c$ is the gradient of the task loss with respect to this activation. This score approximates the change in task loss if the channel is removed.

In a multi-task setting, we compute importance scores for object detection, drivable area segmentation, and lane line segmentation using the Taylor-based criterion introduced above. Each task score $I_c^t$ is first normalized individually to the range $[0,1]$ to make across tasks comparable. To obtain a unified score across tasks, we aggregate the per-task values using a softmin operator:
\begin{equation}
I_c = \operatorname{Softmin}_\tau\big(I_c^{\text{det}}, \; I_c^{\text{da}}, \; I_c^{\text{lane}}\big),
\end{equation}
where $\tau$ is a temperature parameter that controls the sharpness of the aggregation. The softmin operator is defined as
\begin{equation}
\operatorname{Softmin}_\tau(\{I_c^t\}) = -\tau \cdot \log \sum_{t} \exp \left(-\frac{I_c^t}{\tau}\right).
\end{equation}
This process can find out the weaker task-specific importance. Then, set the high priority to delete when the channel is not important across all tasks. 

\subsubsection{Gradient Conflict Penalty}
Although task importance estimation identifies channels that are critical for individual tasks, it does not capture whether a channel introduces interference across tasks. In multi-task learning, the gradients from different tasks may conflict with each other, causing certain channels to hinder optimization. To address this, we introduce a channel-level gradient conflict penalty. 

Formally, let $g_c^t$ denote the average gradient of channel $c$ for task $t$. 
For a task pair $(t,t')$, we measure their per-channel gradient alignment using a stabilized sign-based similarity:
\begin{equation}
\text{sim}(g_c^t, g_c^{t'}) = 
\frac{g_c^t \, g_c^{t'}}{|g_c^t| \, |g_c^{t'}| + \varepsilon},
\end{equation}
where $\varepsilon$ is a small constant that prevents division by zero and is also considered as a soft threshold to suppress spurious conflicts from near-zero gradients. When both gradients have sufficiently large magnitude and the same sign, the similarity is close to $+1$. If they have opposite signs, the similarity is close to $-1$. When at least one gradient is negligible relative to $\varepsilon$, the similarity approaches $0$. We then compute the minimum similarity across all task pairs and define the channel-level conflict penalty as:
\begin{equation}
P_c = \max \Big( 0,\; - \min_{t \neq t'} \text{sim}(g_c^t, g_c^{t'}) \Big).
\end{equation}
A larger $P_c$ quantifies stronger inter-task gradient conflict on channel $c$.

\subsubsection{Safe Pruning Mechanism}
Although the aggregated channel importance $I_c$ and channel gradient conflict penalty $P_c$ provide information for pruning, directly ranking channels based on these values risks removing an important channel. For example, a channel that is critical for one task but conflicting across tasks could be mistakenly pruned. To alleviate such risk, we introduce a safe pruning mechanism. 

For each channel $c$, we compute two statistics from the task-normalized importance scores: 
the maximum importance $I_{\max}(c)$ and the average importance $I_{\text{avg}}(c)$. 
A channel is considered safe to prune only if
\begin{equation}
\text{safe}_c =
\big(I_{\max}(c) < \theta_{\text{max}}\big)\ \land\
\big(I_{\text{avg}}(c) < \theta_{\text{avg}} \ \ \text{or}\ \ P_c > \theta_{\text{pen}}\big),
\end{equation}
where $\theta_{\text{max}},\theta_{\text{avg}},\theta_{\text{pen}}$ are thresholds.

Given this condition, the final pruning score is defined as
\begin{equation}
S_c =
\begin{cases}
I_c - \lambda\, P_c, & \text{if } \text{safe}_c \text{ is true},\\[2pt]
+\infty, & \text{otherwise},
\end{cases}
\end{equation}
where $\lambda$ is a weight of the conflict penalty. We set a global pruning rate $r$ for each eligible backbone layer. Specifically, the channels in each layer are ranked in ascending order of $S_c$, and the bottom $r$ fraction is removed. For structural safety, channels marked as unsafe are assigned $+\infty$ and are thus excluded from pruning. The number of remaining channels in each layer is further aligned to multiples of 8 to ensure hardware efficiency. Pruning focuses on the backbone convolutional layers, while task heads, adapters, and attention modules are excluded.

\subsection{Knowledge Distillation}
To recover accuracy after pruning, we introduce a feature-level knowledge distillation strategy. Instead of aligning task-specific logits, we distill intermediate backbone and encoder features, making the method needn't consider task head and computationally efficient. Additionally, this design is particularly beneficial for multi-task learning, where task heads often differ in both number and structure. As a result, our strategy is more general and easier to extend. Algorithm \ref{alg:kd} illustrates our knowledge distillation process.

We use the RMT-PPAD proposed in \cite{wang2025rmt} as the teacher model, which is already well-trained on BDD100k. During the knowledge distillation, the teacher runs in reduced precision (fp16) to reduce the GPU memory occupation and freezes the weights. 

The overall training loss is defined as:
\begin{equation}
\mathcal{L}_{\mathrm{total}} = \sum_{t \in \mathcal{T}} \mathcal{L}_t + \beta \,\mathcal{L}_{\mathrm{KD}},
\end{equation}
where $\mathcal{L}_t$ is the task-specific loss followed \cite{wang2025rmt}, $\mathcal{T}$ is the set of three tasks, $\beta$ is the knowledge distillation loss weight, and $\mathcal{L}_{\mathrm{KD}}$ is the feature-level knowledge distillation loss. The KD loss is defined as:
\begin{equation}
\label{eq:kd}
\mathcal{L}_{\mathrm{KD}} = 
\frac{1}{|\mathcal{K}|}
\sum_{\ell \in \mathcal{K}}
MSE(P_s^\ell(\mathbf{S}_\ell), P_t^\ell(\widehat{\mathbf{T}}_\ell)),
\end{equation}
where $\mathcal{K}$ is the set of selected layers for distillation, $\mathbf{S}_\ell \in \mathbb{R}^{B \times C_s^\ell \times H_s^\ell \times W_s^\ell}$ and $\mathbf{T}_\ell \in \mathbb{R}^{B \times C_t^\ell \times H_t^\ell \times W_t^\ell}$ denote the student and teacher features at layer $\ell$. Here $\ell \in \mathcal{K}$. We first spatially align teacher features to the student resolution by bilinear interpolation:
$\widehat{\mathbf{T}}_\ell = resize(\mathbf{T}_\ell;\; H_s^\ell,\, W_s^\ell)$
Then, both features are projected to the same dimension via $1{\times}1$ linear projections $P_s^\ell$ and $P_t^\ell$.

\begin{algorithm}[h]
\caption{Feature Knowledge Distillation}
\label{alg:kd}
\begin{algorithmic}[1]
\Require Image batch $\mathbf{X}$; student $\mathcal{M}_S$; teacher $\mathcal{M}_T$;
distillation layer set $\mathcal{K}$; knowledge distillation weight $\beta$
\Ensure Total loss $\mathcal{L}_{\mathrm{total}}$
\State Freeze all parameters of $\mathcal{M}_T$
\State Convert $\mathcal{M}_T$ to the fp16 dtype
\State Set $\mathcal{M}_T$ to evaluation mode
\State Run $\mathcal{M}_S(\mathbf{X})$ to obtain predictions and student features $\{\mathbf{S}_\ell\}_{\ell\in\mathcal{K}}$
\State Compute task losses $\{\mathcal{L}_t\}_{t\in\mathcal{T}}$
\State Run $\mathcal{M}_T(\mathbf{X})$ to collect teacher features $\{\mathbf{T}_\ell\}_{\ell\in\mathcal{K}}$
\For{each $\ell \in \mathcal{K}$}
  \State $\widehat{\mathbf{T}}_\ell \gets resize(\mathbf{T}_\ell;\; H_s^\ell, W_s^\ell)$ 
  \State $\mathbf{S}'_\ell \gets P_s^\ell(\mathbf{S}_\ell)$,\quad $\mathbf{T}'_\ell \gets P_t^\ell(\widehat{\mathbf{T}}_\ell)$
\EndFor
\State $\displaystyle \mathcal{L}_{\mathrm{KD}} \gets \frac{1}{|\mathcal{K}|}\sum_{\ell\in\mathcal{K}} MSE(\mathbf{S}'_\ell,\mathbf{T}'_\ell)$
\State $\displaystyle \mathcal{L}_{\mathrm{total}} \gets \sum_{t\in\mathcal{T}}\mathcal{L}_t \;+\; \beta\,\mathcal{L}_{\mathrm{KD}}$
\State \Return $\mathcal{L}_{\mathrm{total}}$
\end{algorithmic}
\end{algorithm}

\begin{figure*}[h]
  \centering
  \begin{tabular}{l @{\quad} cccc}
    \rotatebox{90}{\textbf{RMT-PPAD}} &
      \includegraphics[width=0.22\linewidth]{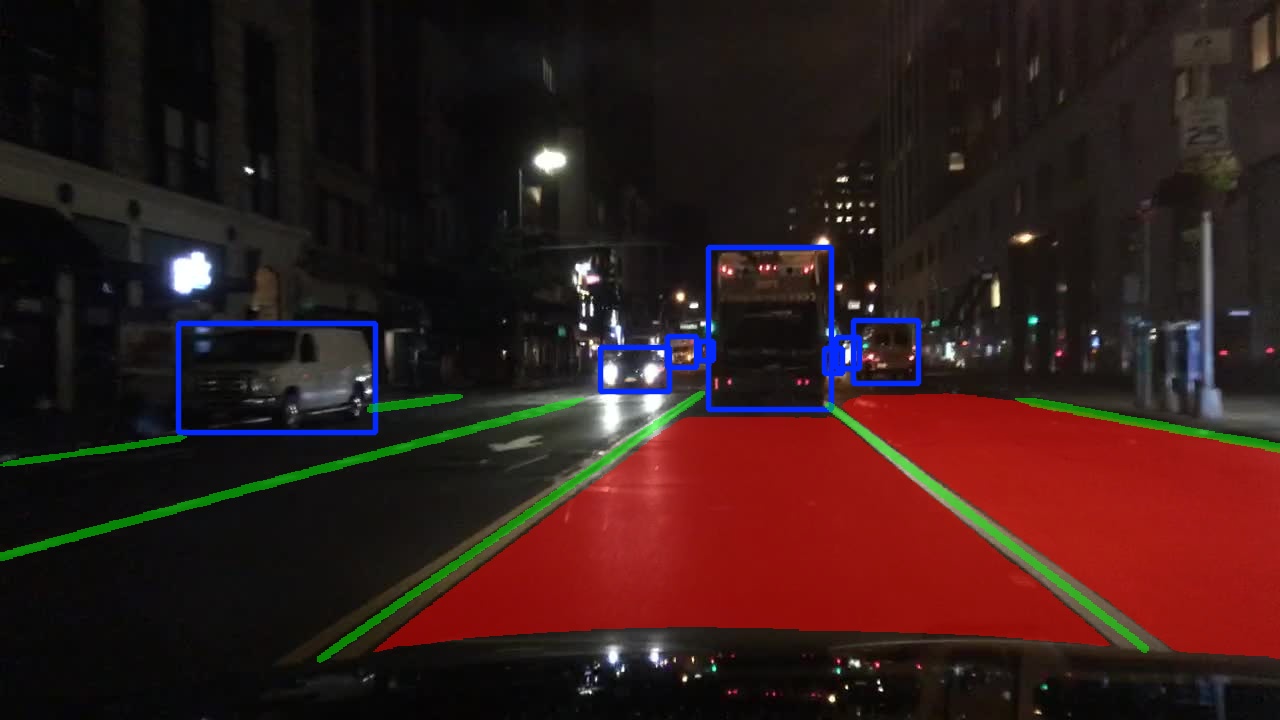} &
      \includegraphics[width=0.22\linewidth]{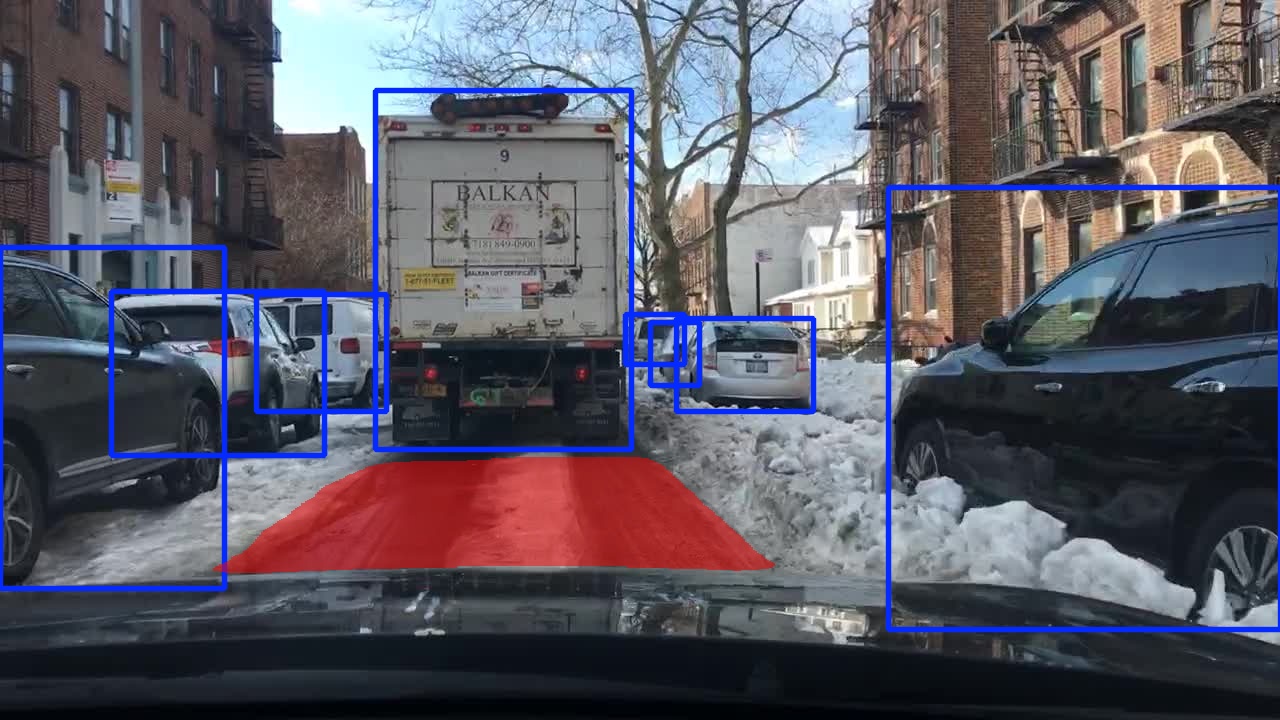} &
      \includegraphics[width=0.22\linewidth]{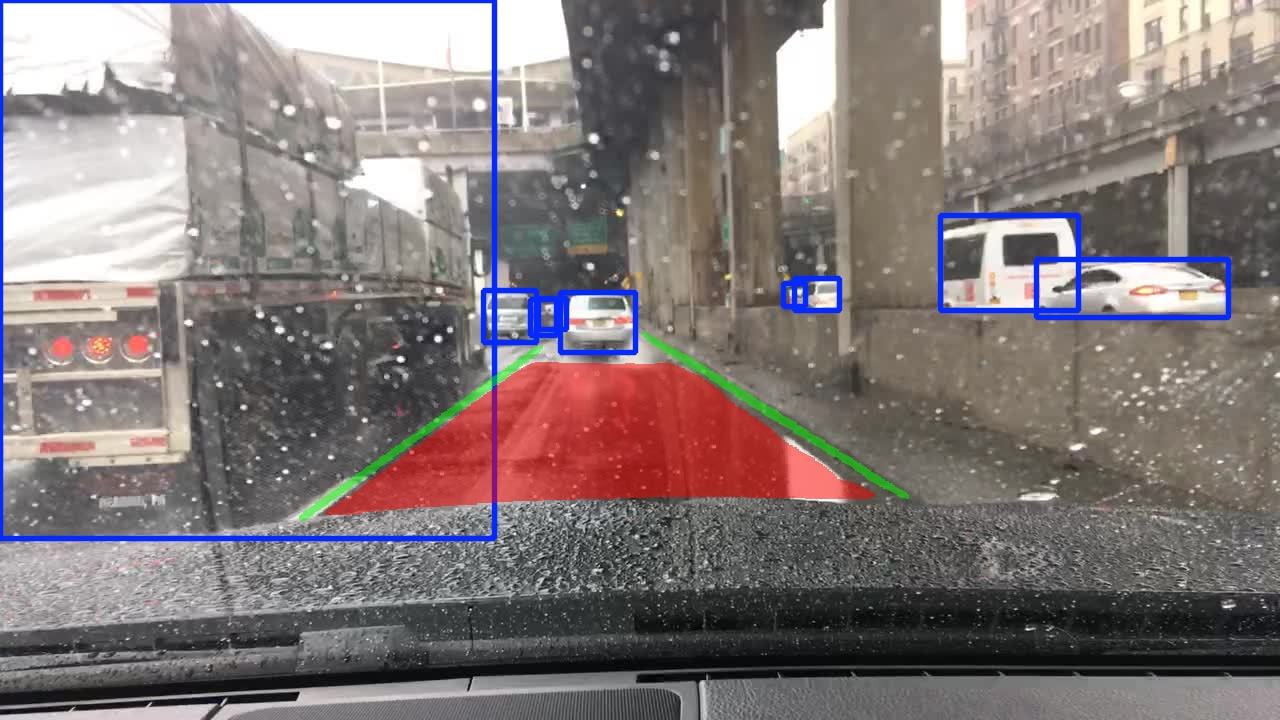} &
      \includegraphics[width=0.22\linewidth]{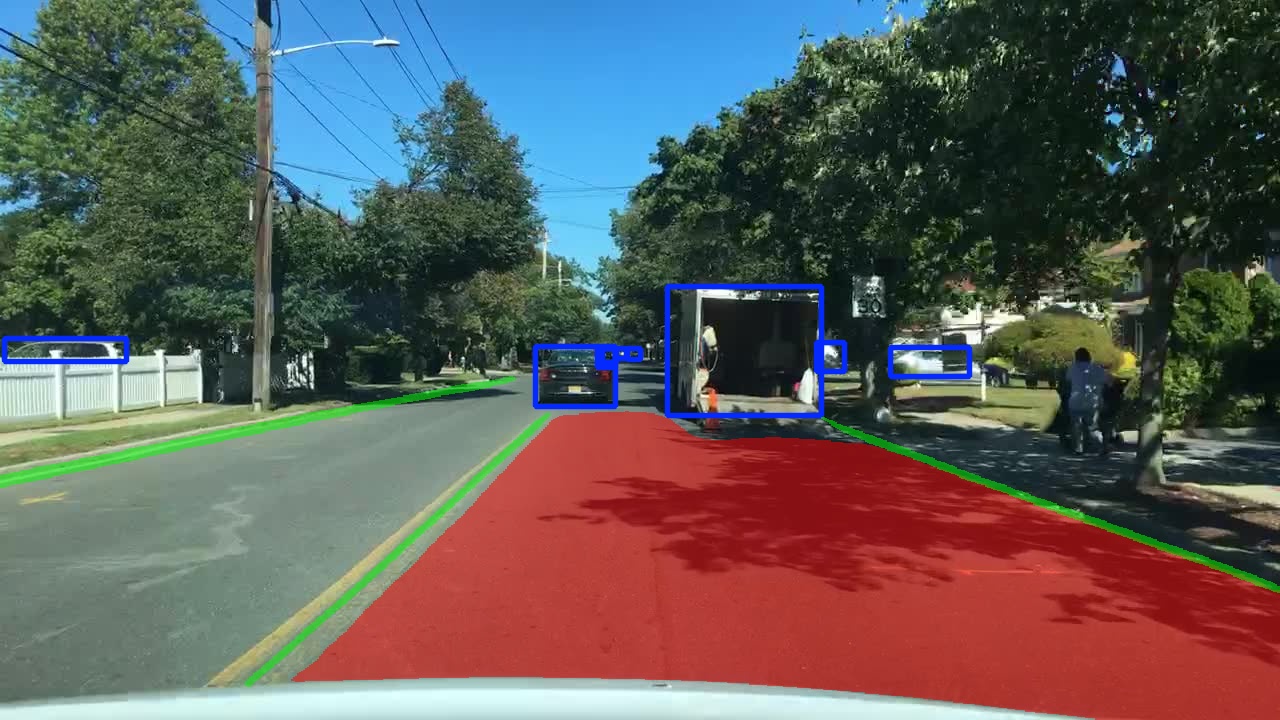} \\[6pt]

    \rotatebox{90}{\textbf{Ours}} &
      \includegraphics[width=0.22\linewidth]{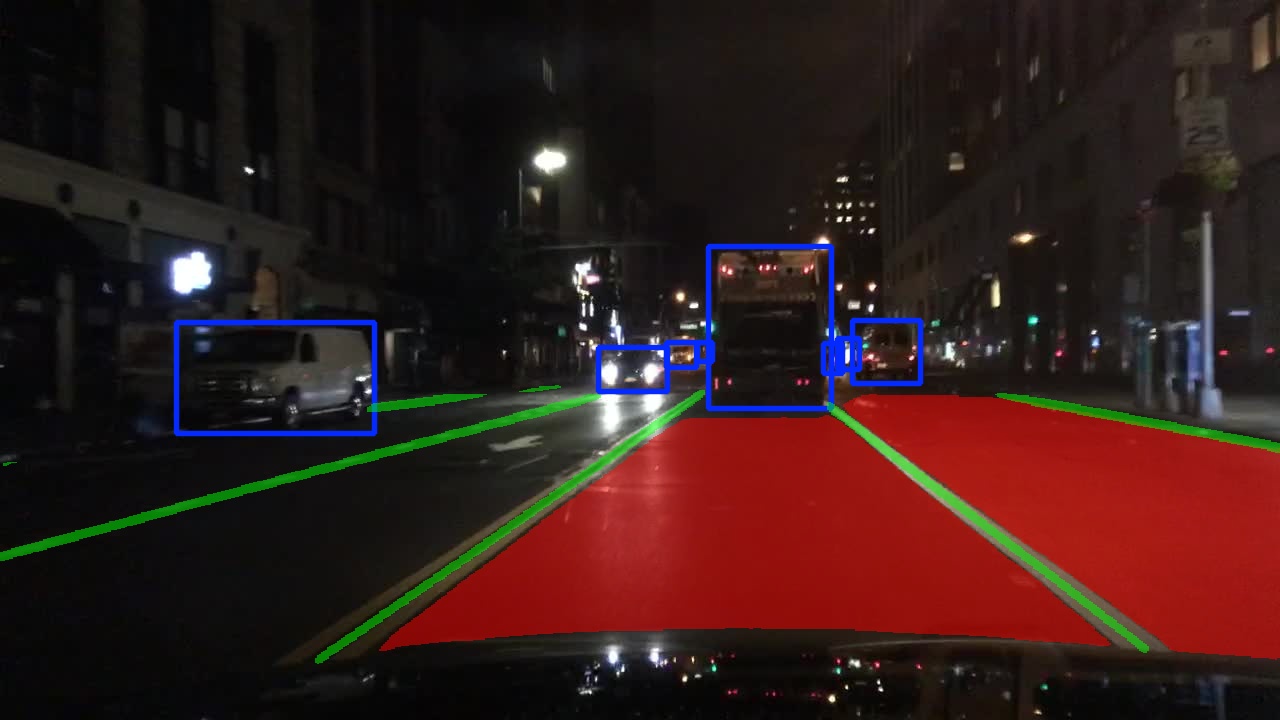} &
      \includegraphics[width=0.22\linewidth]{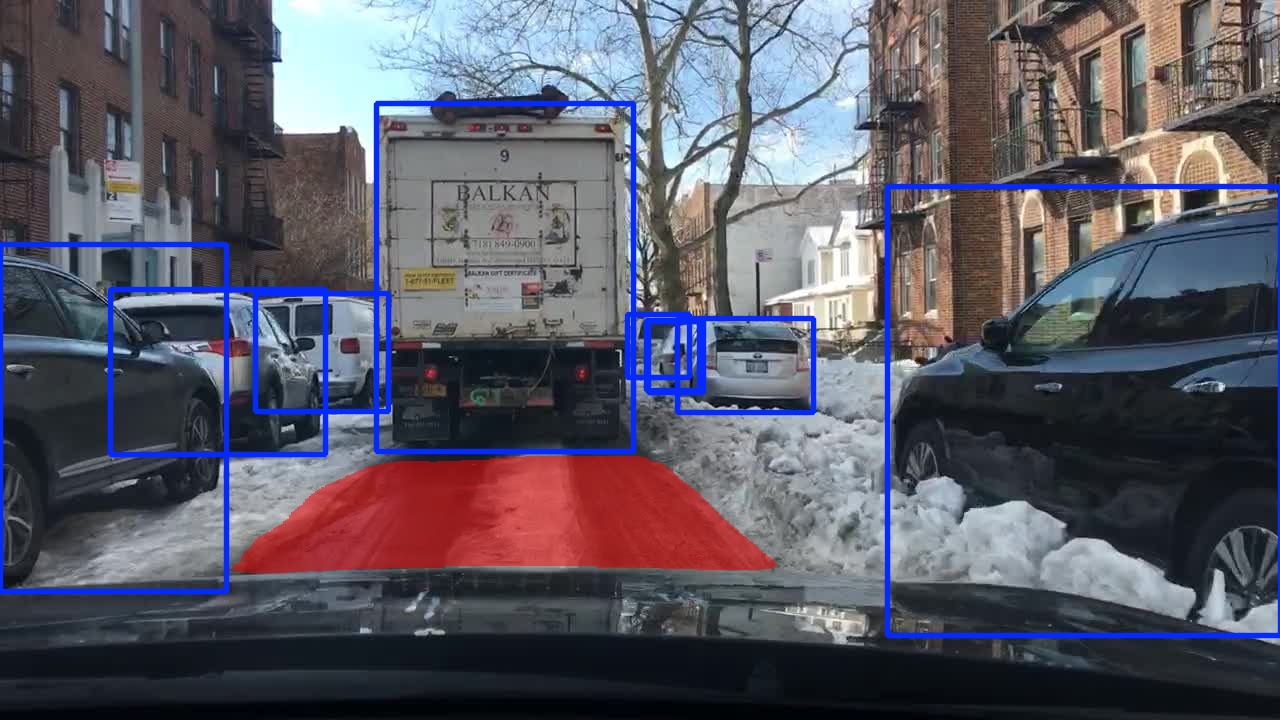} &
      \includegraphics[width=0.22\linewidth]{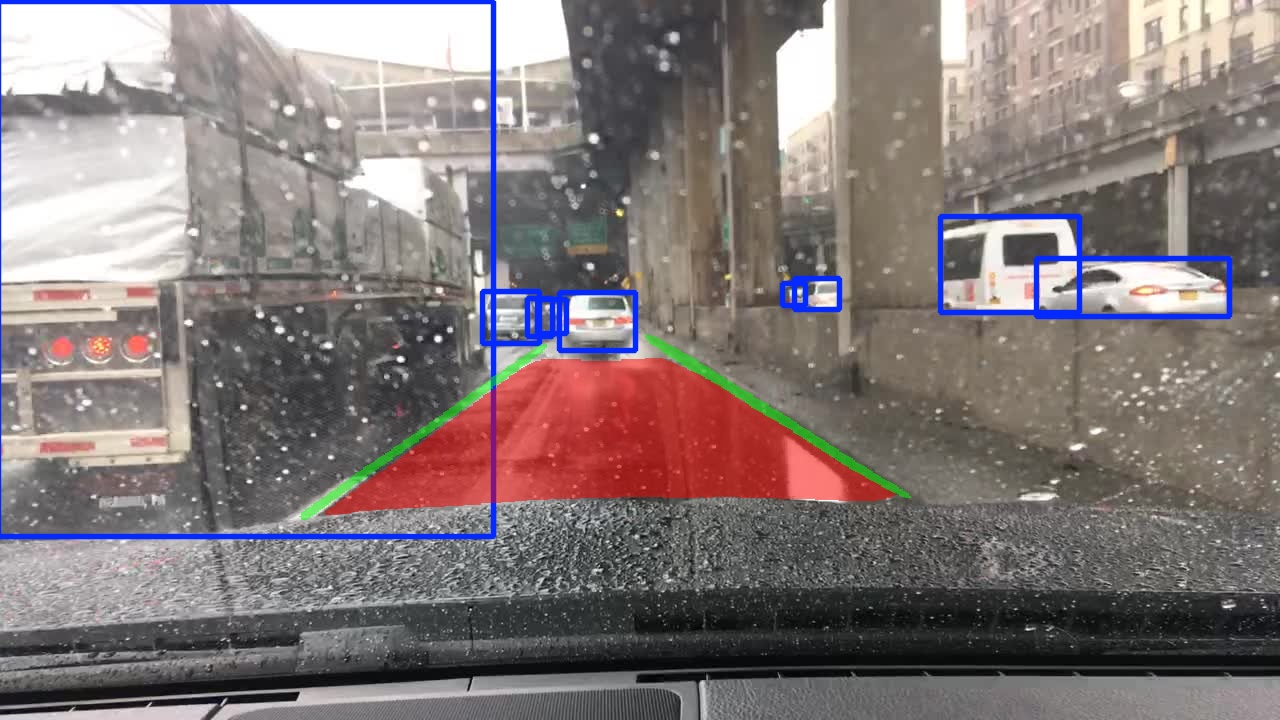} &
      \includegraphics[width=0.22\linewidth]{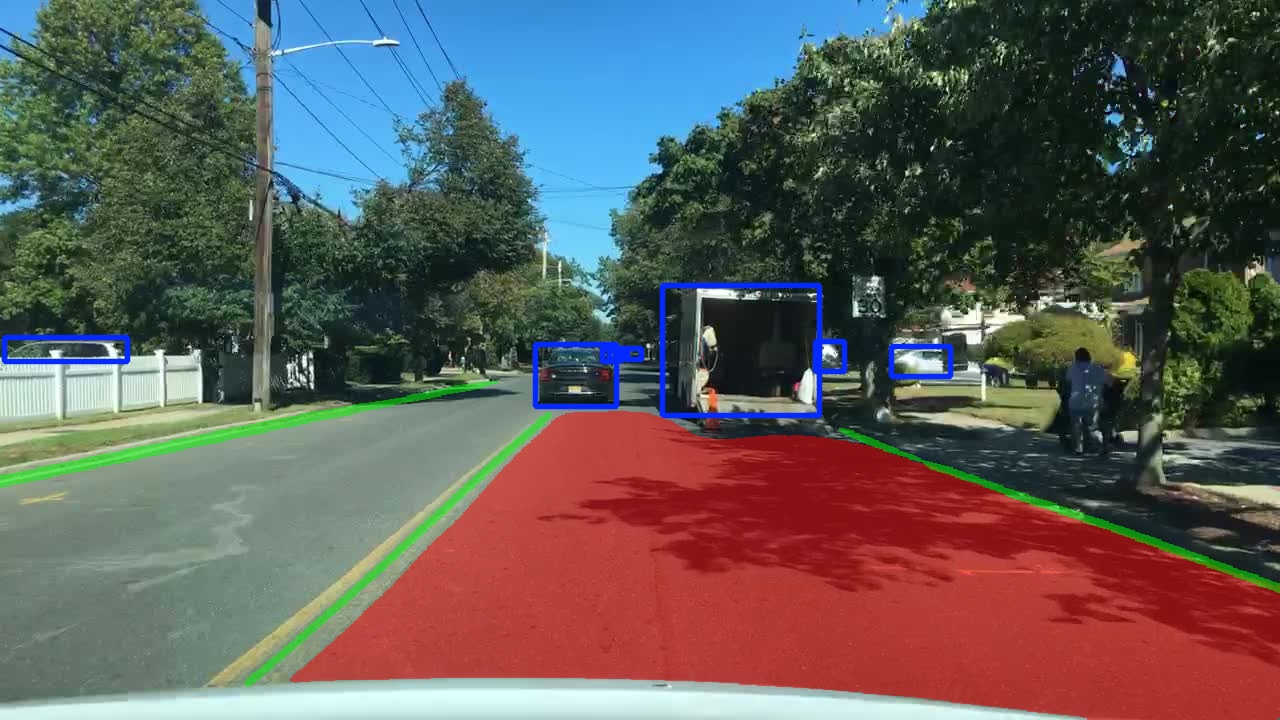} \\
  \end{tabular}
  \caption{Visualization results comparison on different scenarios. From left to right are night, snow, rain, and daytime. }
  \label{fig:real_road}
\end{figure*}

\begin{table*}
\centering
\caption{Quantitative results for each component of the ablation studies on BDD100K.}
\label{tab:comparison_bdd100k}
\renewcommand{\arraystretch}{1.2}
\begin{tabularx}{\textwidth}{c c c *{6}{>{\centering\arraybackslash}X}}
\toprule
\multirow{2}{*}{\textbf{Model}} & \multirow{2}{*}{\textbf{FPS}} & \multirow{2}{*}{\textbf{Params (M)}} &
\multicolumn{2}{c}{\textbf{Object Detection}} &
\multicolumn{1}{c}{\textbf{Drivable Area}} &
\multicolumn{2}{c}{\textbf{Lane Line}} \\
\cmidrule(lr){4-5} \cmidrule(lr){6-6} \cmidrule(lr){7-8}
& & & \textbf{Recall (\%)} & \textbf{mAP50 (\%)} & \textbf{mIoU (\%)} & \textbf{IoU (\%)} & \textbf{ACC (\%)} \\
\midrule
YOLOP       & 64.5  & 7.9  & 88.5 & 76.4 & 89.0  & 44.0 & 79.8 \\
HybridNet   & 17.2  & 12.8 & 93.5 & 77.2 & 91.0  & 52.0 & 82.7 \\
YOLOPX      & 27.5  & 32.9 & 93.7 & 83.3 & 90.9  & 52.1 & 79.1 \\
A-YOLOM(n)  & 52.9 & 4.4  & 85.3 & 78.0 & 90.5  & 45.6 & 77.2 \\
A-YOLOM(s)  & 52.7  & 13.6 & 86.9 & 81.1 & 91.0  & 49.7 & 80.7 \\
RMT-PPAD        & 32.6  & 34.3 & 95.4 & 84.9 & 92.6  & 56.8 & 84.7 \\
Ours        & 32.7  & 23.1 & 94.2 & 83.1 & 92.5  & 56.4 & 85.5 \\
\bottomrule
\end{tabularx}
\end{table*}

\section{Experiments}
In this section, we test our compressed model on the BDD100k \cite{yu2020bdd100k} dataset, which includes multi-task autonomous driving perception scenarios. Additionally, we present the ablation studies to evaluate our proposed pruning and knowledge distillation. 

\subsection{Experiment Details}
For the dataset and each task evaluation metrics, we follow RMT-PPAD \cite{wang2025rmt} to set. Specifically, we based on the RMT-PPAD to compress and evaluate, which can implement three tasks simultaneously: object detection, drivable area segmentation, and lane line segmentation. The evaluation metrics of object detection are mAP50 and Recall. For drivable segmentation, the evaluation metric is mean intersection over union (mIoU). For lane line segmentation, the evaluation metrics are intersection over union (IoU) and pixel accuracy (ACC). We use the complete BDD100k \cite{yu2020bdd100k} dataset to compare our methods with other state-of-the-art multi-task learning methods and use a toy BDD100k dataset for ablation studies. 

Here are some settings about pruning and knowledge distillation. The temperature parameter in softmin aggregation is $\tau=0.25$. $\varepsilon$ in similarity is 1e-12. The thresholds in the safe pruning mechanism are $\theta_{\text{max}}=0.2$, $\theta_{\text{avg}}=0.2$, and $\theta_{\text{pen}}=0.3$. The weight of the conflict penalty in the pruning score is $\lambda=0.2$. The global pruning rate $r=0.4$. The knowledge distillation loss weight $\beta=1$. The selected layers set for distillation $\mathcal{K}$ is ['model.9.ec.conv', 'model.10.conv', 'model.16.cv3', 'model.21.cv3', 'model.24.cv3', 'model.27.cv3']. 

For the distillation training, we set the batch size to 36 for 125 epochs on 3 NVIDIA RTX 4090 GPUs. Additionally, we warm up for 5 epochs, and after that, train the student with knowledge distillation. The FPS evaluation is performed on one NVIDIA RTX 4090 GPU with 1 batch size. For the ablation studies, we decrease the training epoch to 100 because the toy BDD100k dataset only has 10K training samples \cite{wang2025rmt}. Other settings are the same.

\subsection{Experimental Results}

\subsubsection{Quantitative results}

Table \ref{tab:comparison_bdd100k} shows the quantitative results on the BDD100K dataset, comparing our compressed model with several state-of-the-art multi-task learning models, including YOLOP \cite{wu2022yolop}, HybridNet \cite{vu2022hybridnets}, YOLOPX \cite{zhan2024yolopx}, A-YOLOM \cite{wang2024you}, and RMT-PPAD \cite{wang2025rmt}. We evaluate three perception tasks: object detection, drivable area segmentation, and lane line segmentation.

From the results, our compressed model (ours) achieves a balance between accuracy and efficiency. In object detection, our model achieves a recall of 94.2\% and an mAP50 of 83.1\%, which are slightly lower than the teacher model (RMT-PPAD), which is 95.4\% and 84.9\%. However, our model significantly surpasses YOLOP, HybridNet, A-YOLOM(n) and A-YOLOM(s). For drivable area segmentation, our method reaches an mIoU of 92.5\%, close to the best performance achieved by the teacher's 92.6\%. In lane line segmentation, our approach achieves an IoU of 56.4\%, slightly lower than the teacher's 56.8\%, but reaches the best pixel accuracy of 85.5\%, surpassing the teacher’s 84.7\%.

It is noteworthy that our compressed model reduces the parameters to 23.1M. The parameters reduction of approximately 32.7\% compared to RMT-PPAD (34.3M), while only the detection task performance decreases slightly, other segmentation performances are close. Moreover, our compressed model achieves 32.7 FPS for real-time inference.

These results demonstrate the effectiveness of our proposed safe pruning and feature-level knowledge distillation framework. Our method not only significantly reduces parameters but also maintains high accuracy. Deploying models with small parameters has lower memory usage and computing resource requirements, which is crucial for resource-limited or on-board platforms. This makes our compressed model highly suitable for deployment in real-time autonomous driving systems. 

\subsubsection{Visual Results}

Figure \ref{fig:real_road} shows visualization results between our compressed model and the teacher model under different autonomous driving scenarios, including night, snow, rain, and daytime. Overall, both models produce highly consistent predictions in terms of drivable area segmentation. For lane line segmentation, our method achieves results comparable to the teacher model in most scenarios. However, under the night scenario, our compressed model misses a lane line compared to RMT-PPAD. For the object detection, only the bounding box size of the truck under snow scenarios does not match the teacher. This discrepancy is minor, but it still affects the evaluation metric. These visualizations validate that our compressed model maintains strong performance and matches the quantitative results.

\subsection{Ablation Studies}
\begin{table}
  \centering
  \caption{Ablation study for each proposed component on BDD100k toy dataset.}
  \label{tab:module_abaltion}
  \resizebox{\columnwidth}{!}{
  \begin{tabular}{@{} l c c c c c @{}}
    \toprule
    Method                & Recall (\%) & mAP50 (\%) & mIoU (\%) & IoU (\%) & ACC (\%) \\
    \midrule
    RMT-PPAD           & 92.1        & 78.3       & 91.3      & 52.7     & 84.1     \\
    +TCI     & 82.5           & 65.2          & 91.1      & 52.4        & 83.4        \\
    +TCI+GCP         & 86.5           & 72.1          & 91.3         & 52.6     & 83.4     \\
    +TCI+GCP+KD      & 91.8           & 77.7          & 91.2      & 52.4     & 83.3     \\
    \bottomrule
  \end{tabular}
  }
\end{table}

Table~\ref{tab:module_abaltion} presents the ablation study results on the BDD100K toy dataset. These results evaluate the effectiveness of each proposed component: Taylor-based Channel Importance (TCI), Gradient Conflict Penalty (GCP), and Knowledge Distillation (KD). 

We set the teacher model (RMT-PPAD) as the baseline. Directly applying the TCI pruning result to a sharp decrease in detection performance. Specifically, Recall drops from 92.1\% to 82.5\% and mAP50 from 78.3\% to 65.2\%. The reason for this result may removes unimportant channels while enhancing the impact of gradient conflict channels on the overall performance. After GCP is added (+TCI+GCP), the performance partially recovers. Recall achieves 86.5\% and mAP50 achieves 72.1\%, showing that the gradient conflict penalty helps remove the strong gradient conflict channel for multi-task optimization. Finally, we introduce a feature-level KD (+TCI+GCP+KD) further recovers the performance, with Recall reaching 91.8\% and mAP50 improving to 77.7\%. This result is very close to the baseline. For segmentation tasks, the mIoU, IoU, and ACC remain stable across all experiments. This demonstrates that our pruning and distillation strategies primarily affect the detection task in multi-task learning. 

These ablation study results validate the effectiveness of our proposed framework combining TCI, GCP, and KD. Specifically, adding the gradient conflict penalty into pruning is helpful for multi-task optimization, and knowledge distillation compensates for accuracy loss, leading to a compressed model that balances performance across all tasks.

\section{Conclusion}
In this work, we propose a compression framework for multi-task autonomous driving perception models by combining safe pruning and feature-level knowledge distillation. The Taylor-based channel importance and gradient conflict penalty can provide information to perform pruning. This information accounts for channel importance and both inter-task relevance and interference. The feature-level knowledge distillation strategy further ensures that the pruned student model recovers the performance. Experiments on the BDD100K validated that the proposed compression framework reduces model parameters significantly while preserving strong performance across multiple perception tasks. Specifically, compared to the teacher, student segmentation performance is close, and detection performance experiences only a minor decrease. These results confirm that our framework is effective. Future work will focus on extending the framework to other multi-task driving perception tasks and exploring quantization technology to further accelerate the inference speed.






%
\bibliographystyle{IEEEtran}

\bibliography{ref}

\end{document}